\documentclass{article}
\usepackage{ijcai22}

\usepackage{times}
\usepackage{soul}
\usepackage{url}
\usepackage[hidelinks]{hyperref}
\usepackage[utf8]{inputenc}
\usepackage[small]{caption}
\usepackage{graphicx}
\usepackage{amsmath}
\usepackage{amsthm}
\usepackage{booktabs}
\usepackage{algorithm}
\usepackage{algorithmic}
\usepackage{color}
\title{MAP-SNN: Mapping Spike Activities with Multiplicity, Adaptability, and Plasticity into Bio-Plausible Spiking Neural Networks}

\author{
Chengting Yu$^{1,2}$
\and
Yangkai Du$^3$\and
Mufeng Chen$^1$\and
Aili Wang$^{1,2,*}$\and
Gaoang Wang$^2$\And
Erping Li$^{1,2}$
\affiliations
$^1$College of Information Science and Electronic Engineering, Zhejiang University\\
$^2$ZJU-UIUC Institute, Zhejiang University\\
$^3$College of Computer Science and Technology, Zhejiang University
\emails
chengting.21@intl.zju.edu.cn,
\{yangkaidu, chenmufeng\}@zju.edu.cn,
ailiwang@intl.zju.edu.cn
}

\begin{document}

\maketitle

\begin{abstract}
Spiking Neural Network (SNN) is considered more biologically realistic and power-efficient as it imitates the fundamental mechanism of the human brain.
Recently, backpropagation (BP) based SNN learning algorithms that utilize deep learning frameworks have achieved good performance.
However, bio-interpretability is partially neglected in those BP-based algorithms.
Toward bio-plausible BP-based SNNs, we consider three properties in modeling spike activities: Multiplicity, Adaptability, and Plasticity (MAP).
In terms of multiplicity, we propose a Multiple-Spike Pattern (MSP) with multiple spike transmission to strengthen model robustness in discrete time-iteration. 
To realize adaptability, we adopt Spike Frequency Adaption (SFA) under MSP to decrease spike activities for improved efficiency.
For plasticity, we propose a trainable convolutional synapse that models spike response current to enhance the diversity of spiking neurons for temporal feature extraction.
The proposed SNN model achieves competitive performances on neuromorphic datasets: N-MNIST and SHD. 
Furthermore, experimental results demonstrate that the proposed three aspects are significant to iterative robustness, spike efficiency, and temporal feature extraction capability of spike activities.
In summary, this work proposes a feasible scheme for bio-inspired spike activities with MAP, offering a new neuromorphic perspective to embed biological characteristics into spiking neural networks.


\end{abstract}


\section{Introduction}

Motivated by biological plausibility, Spiking Neural Network (SNN) is introduced as a noise-robust third-generation neural network \cite{maass_networks_1997}. The SNN transmits the discrete action potentials (spikes) through the adaptive synapses to process information similar to the communication scheme in the brain. Therefore, the exploration of SNN is anticipated to help reveal the working mechanism of the mind and intelligence \cite{ghosh-dastidar_spiking_2009}. 
Besides, the event-driven characteristic of SNN allows it to be potentially energy-efficient on the emerging neuromorphic hardware and relatively mature neuromorphic sensors \cite{vanarse_review_2016}.

However, designs and analyses of SNN training algorithms are challenging.
The asynchronous and discrete computing in SNN makes it difficult to apply the mature backpropagation (BP) technique for practical training \cite{pfeiffer_deep_2018}.
A pseudo-derivative method is introduced in recent work to overcome the non-differentiable problem, allowing SNN to be directly trained using BP \cite{wu_spatio-temporal_2018}.
Those BP-based SNNs utilize the basic concept of Recurrent Neural Network (RNN) by converting spiking neurons into an iterative model and simulating neural activities with discrete time-iteration.
With BP-based learning algorithms, SNN models can be implemented on a larger scale under mature deep learning frameworks to achieve better performances \cite{wu_direct_2019,wozniak_deep_2020}. 


At present, some bio-inspired SNNs reveal the potential of biological characteristics with better performance, such as Lateral Interactions \cite{cheng_lisnn_2020}.
Inspired by neuroscience, this work focuses on the neuromorphic properties of spike activities and proposes a feasible bio-plausible scheme with Multiple-Spike Pattern (MSP), Spike Frequency Adaption (SFA), and Convolutional Synapse (ConvSyn), advancing BP-based SNN toward the goal of neuromorphic computing.
The multiple-spike pattern for spiking neurons allows multiple spikes transmission at the minimal iterative step length in terms of multiplicity. Furthermore, we adopt the Spike Frequency Adaptation (SFA) mechanism under MSP to realize adaptability for higher efficiency of spike activities. Compared with the single-spike pattern, the proposed multiple-spike pattern with the SFA mechanism alleviates the problem caused by discrete time-iteration and results in better model stability under different step lengths.
Besides, inspired by the synaptic plasticity, this work proposes a convolutional synapse model to imitate the bio-electric synapse for converting incoming spike trains into pre-synaptic currents, further enhancing the temporal extraction ability of spike activities. 

We test the proposed model on two neuromorphic datasets: N-MNIST and SHD \cite{orchard_converting_2015,cramer_heidelberg_2020}. The experimental results show that the proposed model can achieve competitive performance on both N-MNIST and SHD. Furthermore, comparative and analytical experiments demonstrate that the proposed scheme has more robust model stability under different iterative step lengths, fewer but practical spike activities, and better model performance for temporal feature extraction in neuromorphic tasks.
 
Our main contributions are four-folds:

\begin{enumerate}
\item This work reveals the distinction between the discrete iterative simulation and biological network. To our best knowledge, this study, for the first time, discusses the discretization problem in time-iteration and raises a new question about the model robustness under different iterative step lengths for BP-based SNN algorithms.

\item This work explores the importance of modeling spike activities and shows researchers in neuromorphic computing more possibilities of embedding biological properties into SNNs.

\item This work proposes the Multiple-Spike Pattern for robust iterative training in Spiking Neural Network, providing a potential direction of SNN algorithm developments.

\item This work proposes a Convolutional Synapse, modeling biological properties using mature convolution operations, making the SNN algorithm more compatible with deep learning frameworks.



\end{enumerate}


\section{Related Work}


SNNs are computational models that consist of spiking neurons and interconnecting synapses with adjustable scalar weights. In this section, we discuss spiking neuron models and learning rules of synapses related to our proposed techniques.


\subsection{Spiking Neuron Models}
The spiking neurons receive input temporal signals and generate a spike when the membrane potential reaches a threshold. Multiple neuron models have been proposed to model the neural spike activities, including Rate,
McCulloch and Pitts,
Hodgkin-Huxley,
and FitzHugh-Nagumo
models \cite{gerstner_spiking_2002}. However, those complex neural models with extensive biological details cause high computational costs. Recently, the leaky integrated-and-fire (LIF) model
has drawn much attention. The LIF model captures the intuitive properties of external input accumulating charge across a leaky cell membrane with a clear threshold \cite{tavanaei_deep_2019}. An explicitly iterative version of the LIF model was generally utilized in nowadays deep learning frameworks \cite{wu_spatio-temporal_2018,cheng_lisnn_2020}, allowing discrete neural spike activities in deep SNNs.
(Mathematical derivation of iterative LIF model is included in the supplementary material.)

\begin{figure}[hbtp]
    \centering
    \includegraphics[width = 8cm]{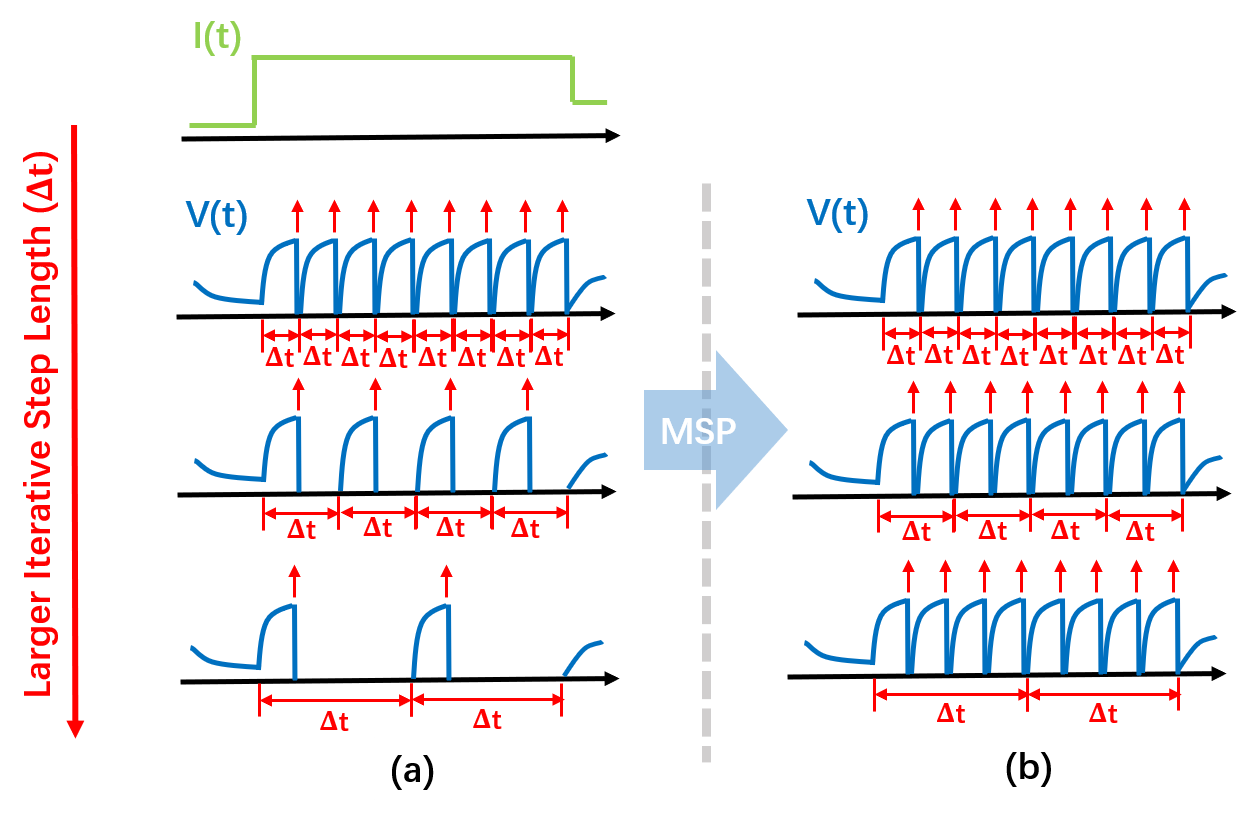}
    \caption{MSP motivation. The discretization of spike activities under time-iteration. $V(t)$ is membrane potential varying with input current $I(t)$. Red arrows are spike activities produced when potential arrives threshold. (a) Modeling Spike activities in Single-Spike Pattern (SSP). Under SSP, spikes are limited when the iterative step length goes larger. (b) Modeling Spike activities Multiple-Spike Pattern (MSP). Under MSP, the spike number is replenished in each iterative step.
    }
    \label{fig:problem}
\end{figure}

\subsection{Learning Rules in SNNs}
The strengths of synapses are modeled as scalar weights in SNNs, which can be dynamically adjusted following a specific learning rule. The learning rules of SNNs are actively explored and can be roughly concluded into three directions: conversion-based methods that map SNNs from trained ANNs \cite{han_rmp-snn_2020}; supervised learning with spikes that directly train SNNs using variations of error backpropagation \cite{lee_training_2016,wu_spatio-temporal_2018}; local learning rules at synapses, such as schemes exploring the spike time dependent plasticity (STDP) \cite{song_competitive_2000}. Recent works have successfully applied the backpropagation algorithm into SNNs by defining pseudo-derivatives for the non-differentiable spike activities \cite{lee_training_2016,wu_spatio-temporal_2018,tavanaei_deep_2019,cheng_lisnn_2020}. 
Those BP-based SNNs are similar to extensions of traditional Recurrent Neural Networks (RNNs), which utilize error backpropagation through time and follow gradient descent to adjust the connection weights.
The BP-based algorithms can take advantage of mature deep learning frameworks for network design and operating efficiency.
Thus, they have become an essential branch of the SNN algorithmic directions.

\subsection{Biological Properties in BP-based SNNs}
As the main target of neuromorphic computing, the research of BP-based SNN combined with biological characteristics is highly concerned. Recently, some works revealed the potential of biological characteristics in SNNs with better performance, such as Lateral Interactions for intra-layer connections \cite{cheng_lisnn_2020}, delayed Spike Response Model (SRM) for synaptic expressions \cite{shrestha_slayer_2018}, providing a good entry point for training BP-based SNNs with bio-interpretability.

\section{Methodology}
\begin{figure*}[htbp]
    \centering
    \includegraphics[width = 17cm]{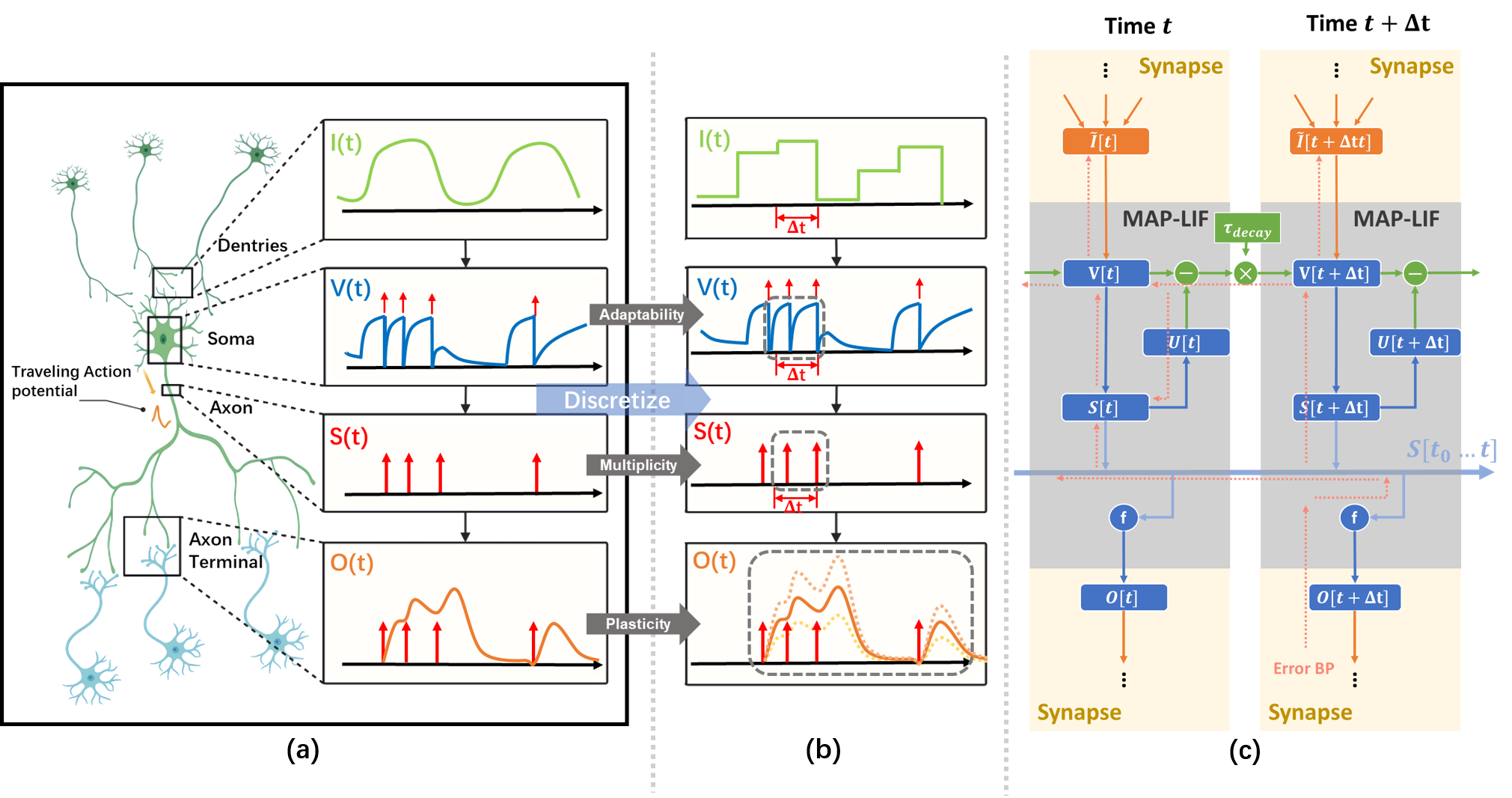}
    \caption{MAP-SNN overview. (a) Modeling neural model mathematically with Dendrites, Soma and Axon. The four variables $I$, $V$, $S$, and $O$  describe the input current, the membrane potential, the spike transmission, and the output current concerning time, respectively. (b) Discretized spike activities with MAP properties. (c) The spatio-temporal feedforward and backward dataflow in the proposed model. Each box of MAP-LIF indicates a certain spatio-temporal iterative state. The solid arrows indicate spatial and temporal feedforward; the red dotted arrows indicate the error backpropagation correspondingly along with the reverse directions of feedforward.
    }
    \label{fig:work_intro}
\end{figure*}

As shown in Figure \ref{fig:work_intro}, we model the spike activities guided by the MAP principles. The implementation of our proposed model with motivation and benefits of each improvement are presented in this section. (The derivation details of formulas can be found in the supplement.)

\subsection{Multiplicity with Multiple-Spike Pattern}
In this part, the discretization problem of iterative models is discussed first. Then, the multiple-spike pattern that represents the model's multiplicity is presented.

\subsubsection{Problem under Discrete-Time-Iteration}
BP-based SNNs simulate the spike activities by discrete time-iteration. However, the discreteness causes considerable problems. Within recursive time-iterations, $\Delta t$ needs to be determined as the minimum iterative step length for simulation. 
The mismatch between the continuous and discrete spike activities is illustrated in Figure \ref{fig:problem}. 
For distinction, we define BP-based SNNs with binary digits transmission as Single-Spike Pattern (SSP). With SSP, modeling spike activities becomes problematic when iterative step length $\Delta t$ goes larger. 
Since models with SSP represent spike activities as binary sequences, only one spike activity can be handled per iterative step. Under this circumstance, the temporal feature is restricted with lost spikes. Therefore, in discrete time-iteration, the proper selection of the iterative step lengths is always an indispensable part of SSP. Inspired by spike multiplicity, 
we propose a Multiple-Spike Pattern (MSP) to ease the problem by complementing spike activities into each iterative step, as shown in Figure \ref{fig:problem}.





\subsubsection{Multiple-Spike Pattern for Discrete Iteration}



\begin{figure}[htbp]
    \centering
    \includegraphics[width = 8.5cm]{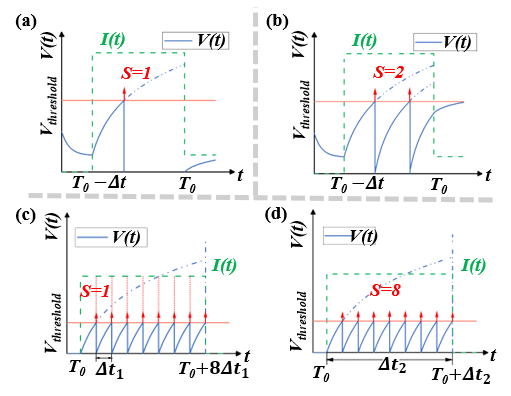}
    \caption{Pattern Comparison. (a) The plot follows the spike activity under a single-spike pattern when an extensive input current $I(t)$ is given. (b) The plot shows how the multiple-spike pattern works to achieve the complete spike activity. (c) Equivalent case of two patterns, when $\Delta t_1$ is applied in SSP and $\Delta t_2=8\Delta t_1$ is applied in MSP, both patterns trigger eight times spike activities during the certain time.}
    \label{fig:Pattern Comparison}
\end{figure}

As the compensation for the difference between neurons' natural behaviors under such discrete iteration, MSP utilizes integers as the intensity of neural spike activities, allowing multiple spikes transmission in an iterative step. The comparison between the SSP and the MSP is shown in Figure \ref{fig:Pattern Comparison}(a) and Figure \ref{fig:Pattern Comparison}(b).
By replenishing the spike number within iterative steps, the multiple-spike pattern avoids the spike loss problem and implements potentially higher expressiveness of spike activities during the time iteration.


\subsubsection{Equivalence under Multiple-Spike Pattern}
The MSP proposed in this work can be equivalently converted into the SSP by reducing the time scale correspondingly. One example is shown in Figure \ref{fig:Pattern Comparison}(c), where two patterns result in the same spike activities during a specific time interval. Furthermore, under the equivalence, the multiple-spike pattern implements the same neuron activity with more leisurely $\Delta t$ selection, which allows the model to achieve high stability for arbitrary iterative step lengths. 



\subsection{Adaptability with Spike Frequency Adaptation}

\subsubsection{Spike Frequency Adaptation for Spike Activities}


Spike-frequency adaptation (SFA) is a biological neural phenomenon, describing a neuron fires with a frequency that reduces over time when stimulated with constant input.
The phenomenon occurs in both vertebrates and invertebrates, in peripheral and central neurons, and plays an essential role in neural information processing \cite{benda_universal_2003}. The SFA mechanism leads to non-linearity in spike activities and enriches the temporal feature for a single spike. 
Specifically, Adibi et al. \shortcite{adibi_informational_2013} suggest that the SFA mechanism in real neurons like whisker sensory cortexes helps improve the information capacity of a single spike defined by the average mutual information (MI).
Therefore, this work adopts the SFA mechanism with MSP for higher efficiency in spike transmission.

\subsubsection{Iterative LIF with SFA}
By unifying the accumulation activity and spike activity, the iterative LIF model with SFA is defined as:
\begin{equation}
     v(t) = \tau_{decay} \times \left\lbrack v\left( {t - \Delta t} \right) - u\left( t - \Delta t \right) \right\rbrack  + \tilde{I}(t)
    \label{eq:SFA}
\end{equation}
Here $v(t)$ is the neuron's membrane potential. $u(t)$ is the consumed membrane potential that produces multiple spike activities. $\tilde{I}(t)$ is the normalized pre-synaptic input current. $\tau_{decay}$ is the decay factor describing the leaky activity of spiking neurons.
\begin{equation}
    n^{*} = {\log_{q}\left\lbrack \frac{v}{V_{threshold}}\left( {q - 1} \right) + 1 \right\rbrack}
\end{equation}

\begin{equation}
    s =  \lfloor n^{*} \rfloor
\end{equation}

\begin{equation}
    u = \sum_{i=1}^{s}{(q^{i}\times V_{threshold})}
    = \frac{q^{s} - 1}{q - 1} \times V_{threshold}
\end{equation}
Here $V_{threshold}$ is the threshold basis  of neuron, $n^*$ is the estimated intensity of spike activity, $s$ is the integer number of output spike activities, $q$ is the inhibition coefficient that controls the temporary raising of the threshold, making the intensity of spike activity drop exponentially. 

\subsubsection{Adaptability with Fewer Spike Activities}
\begin{figure}[htbp]
    \centering
    \includegraphics[width = 8.5cm]{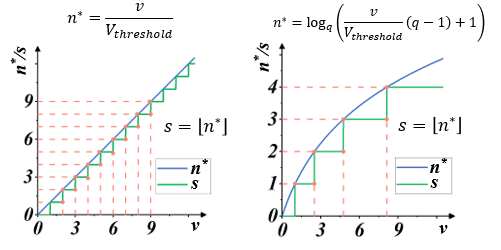}
    \caption{The spike activity versus membrane potential. The blue plot indicates the continuous modeling of the spike intensity, and the green plot indicates the discrete modeling of the spike activities. (a) Spike activities in linear mode. (b) Spike activities in Spike Frequency Adaption mode.}
    \label{fig:s_curve}
\end{figure}

As shown in Figure \ref{fig:s_curve}, when the membrane potential $v$ increases, the intensity of spike activity gradually deviates from linearity, showing adaptability to the current input. In this case, the total number of spike activities decreases, and each spike activity brings more features, potentially saving computation operations with less spike transmission while maintaining high performance.


\subsubsection{Pseudo-Derivative of Spike Activity}
In order to apply backpropagation, we assign a particular pseudo derivative as follows:
\begin{align}
\frac{\partial s}{\partial n^{*}} = 1
\end{align}%
This pseudo-derivative provides a unit vector for gradient descent without complicated computations.


\subsection{Plasticity with Convolutional Synapse}


\subsubsection{Modeling Spike Activity through Electric Synapse}
In biological neural networks with electric synapses, a spike is considered to be generated in a soma, transmitted through the axon to the synapse, and converted as an electric current into the connecting neuron's dendrites. For describing spike transmissions through electric synapses, Gerstner \shortcite{gerstner_spiking_2002} proposes a Spike Response Model (SRM) \ to transform spike activities into current signals flowing into post-synaptic dendrites, defined as:

\begin{equation}
    o(t) = \sum\limits_{i=0}^{i\cdot \Delta t \leq t}s(t-i\cdot \Delta t)K(i\cdot \Delta t)
\end{equation}
Here $s(t)$ is the spike activities, $o(t)$ is the spike response signal transmitted from axon terminal to dendrite over time, $K(t)$ is the spike response kernel relating current intensities with spike activities.



\subsubsection{Potential Plasticity in Spike Response Model (SRM)}
\begin{figure}[htbp]
    \centering
    \includegraphics[width = 5.5cm]{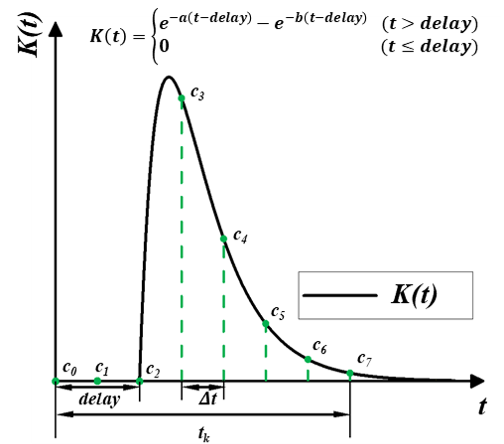}
    \caption{Model the SRM $K(t)$ into 1-D convolution kernel $C_n$. $K(t)$ is modeled with three trainable parameters, with $a$, $b$ for shape, and $delay$ for time delay.}
    \label{fig:SRM}
\end{figure}

SRM provides richer temporal information for the network by allowing the varying effect of certain spike activity. However, the constant parameters of response kernel $K(t)$ are widely pre-defined as a "ground truth" before training, which limits the potential diversity and plasticity for SRM. Shrestha and Orchard \shortcite{shrestha_slayer_2018} first considered the plasticity of SRM by setting response delay as learnable parameters, which unsurprisingly improved the performance. As shown in Figure \ref{fig:SRM}, this work further frees up the shape parameters for better plasticity, allowing shape parameters $a$, $b$, and delay parameter $delay$ to be learnable during training. In this case, the plasticity of spike activity allows each neuron to learn different temporal features, improving the complexity and fitting ability of the model.


\subsubsection{Spike Response Model as 1-D Convolution}
As shown in Figure \ref{fig:SRM}, SRMs used for spike activities are going up to the peak and subsequently decreasing towards zero.
Therefore, it is possible to ignore the long-time spike response to help reduce computational complexity. This work applies one-dimensional convolution operation in computing SRM by defining a valid time window $t_k$ to make the SRM more compatible in nowadays deep learning frameworks. The convolution operation follows as:


\begin{equation}
    o\left( t \right) = f(s,t) = {\sum\limits^{i\cdot \Delta t \leq t_k}_{i=0}{s\left( t-i\cdot \Delta t \right) \times C_i}}
    \label{eq:conv_syn}
\end{equation}

Here $C_n$ is the one-dimensional convolutional kernel of spike responses, modeled with three variables $a$, $b$ and $decay$. $t_k$ is the time-window constant that describes the necessary scope of spike response, defined as Eq.\ref{eq:tk} with the minimal iterative step length $\Delta t$ and convolution kernel size. 
\begin{equation}
    t_{k} = (kernel\_size - 1) \times \mathrm{\Delta}t
    \label{eq:tk}
\end{equation}

\subsubsection{Dendrites with Negative Masking Filter}

In spiking neurons, dendrites receive signals from pre-synaptic axon terminals and integrate all currents together for the neural soma. In this work, we set up a negative filter with $Swish$ function \cite{ramachandran_searching_2018} to shield negative integrated input currents to improve the stability of membrane potential in the LIF model.

\section{Experiment}

\subsection{Experiment Setting}

The proposed model is built on the deep learning framework, PyTorch\footnote{\url{https://pytorch.org/}}, and the weights are initialed using the default xavier\_normal\_\footnote{\url{https://pytorch.org/docs/master/nn.init.html}} method in PyTorch. Besides, we use Adam as the optimizer and Cross-Entropy as the criterion during training. Hyperparameters of experimental settings are included in the supplementary materials with source codes.

To evaluate the performance of the proposed SNN model, we selected two neuromorphic datasets: N-MNIST \cite{orchard_converting_2015} and SHD \cite{cramer_heidelberg_2020}. They are used as experimental objects for classification error rates in neuromorphic tasks, including ablation experiments. In addition, we set up control experiments to analyze and discuss the significance brought from the three characteristics (MAP) to the model performance.

\subsection{Classification on Neuromorphic Datasets}
To clearly demonstrate the reliability of our approaches, we train our SNN models with spike-based datasets for image and sound classification, and compare the achieved error rates with relative works on SNN algorithms.

N-MNIST is a neuromorphic dataset of handwritten digits containing 60,000 train samples and 10,000 test samples. The samples of N-MNIST are event-based spike signals, which are captured by recording digits images on an LCD screen using Dynamic Vision Sensors (DVS). 
Spiking Heidelberg Digits (SHD) is a spike-based speech dataset consisting of 0 to 9 spoken digits recordings in both English and German. The audio recordings are converted into spikes using an artiﬁcial inner ear model, transforming into temporal features with 700 input channels, with 8156 train samples and 2264 test samples.

\begin{table}
\centering
\resizebox{.91\linewidth}{!}{
\begin{tabular}{lcc}
\toprule
Model               &Size of Hidden Layer & Error Rate(\%) \\
\midrule
Spiking-MLP \cite{cohen_skimming_2016}  &10000  &8.13 \\
Spiking-CNN \cite{neil_effective_2016}  &-      &4.28  \\
LSTM \cite{neil_phased_2016}  &-            & 2.95  \\
Phased-LSTM \cite{neil_phased_2016}  &-     & 2.62  \\
MLP \cite{lee_training_2016}    &800        & 2.20  \\
Spiking-MLP \cite{lee_training_2016}         &$800$      & 1.26  \\
STBP \cite{wu_spatio-temporal_2018}  &$800$             & 1.22 \\
Spiking-MLP \cite{fang_neuromorphic_2021}   &$500$-$500$  & 1.60 \\
\textbf{this work (SSP)}   &$800$      & \textbf{1.60} \\
\textbf{this work (ConvSyn)}   &$800$      & \textbf{1.43} \\
\textbf{this work (MSP)}   &$800$      & \textbf{1.11} \\
\textbf{MAP-SNN (MSP+ConvSyn)}  &$800$      & \textbf{1.06} \\
\bottomrule
\end{tabular}}
\caption{Performance of different algorithms on N-MNIST.}
\label{tab:N-MNIST results}
\end{table}

\begin{table}
\centering
\resizebox{.91\linewidth}{!}{
\begin{tabular}{lcc}
\toprule
Model               &Size of Hidden Layer & Error Rate(\%) \\
\midrule

Spiking-MLP \cite{cramer_heidelberg_2020}  &-  &52.5 \\
SNN-base \cite{cramer_heidelberg_2020}  &-  &28.6  \\
R-SNN \cite{cramer_heidelberg_2020}  &-  &16.8 \\
R-SNN \cite{zenke_remarkable_2021}  &-  &18.0  \\
SRNN \cite{yin_effective_2020}  &-  &15.6  \\
Spiking-MLP \cite{fang_neuromorphic_2021}   &$400$-$400$  & 14.3 \\
\textbf{this work (SSP) }  &$400$-$400$      & \textbf{36.1} \\
\textbf{this work (ConvSyn) }  &$400$-$400$      & \textbf{33.0} \\
\textbf{this work (MSP)}   &$400$-$400$     & \textbf{17.1} \\
\textbf{MAP-SNN (MSP+ConvSyn)}   &$400$-$400$     & \textbf{13.0} \\

\bottomrule
\end{tabular}}
\caption{Performance of different algorithms on SHD.}
\label{tab:SHD results}
\end{table}
We compare the obtained optimal model performance with state-of-the-art SNN models, as N-MNIST in Table \ref{tab:N-MNIST results} and SHD in Table \ref{tab:SHD results}, including ablation experiments with MSP and ConvSyn alone. The experimental results show that MAP-SNN can decrease the error rate by $0.2\%$ on N-MNIST and $1.3\%$ on SHD, which has achieved the highest performance among SNN-based algorithms under the same Multilayer Perceptron (MLP) structure. Furthermore, we observe that MSP and ConvSyn are enabled to improve the model accuracy independently and can also be combined together for significantly better performance, which supports the complementarity of MAP properties.

\begin{figure*}[htbp]
    \centering
    \includegraphics[width = 16.5cm]{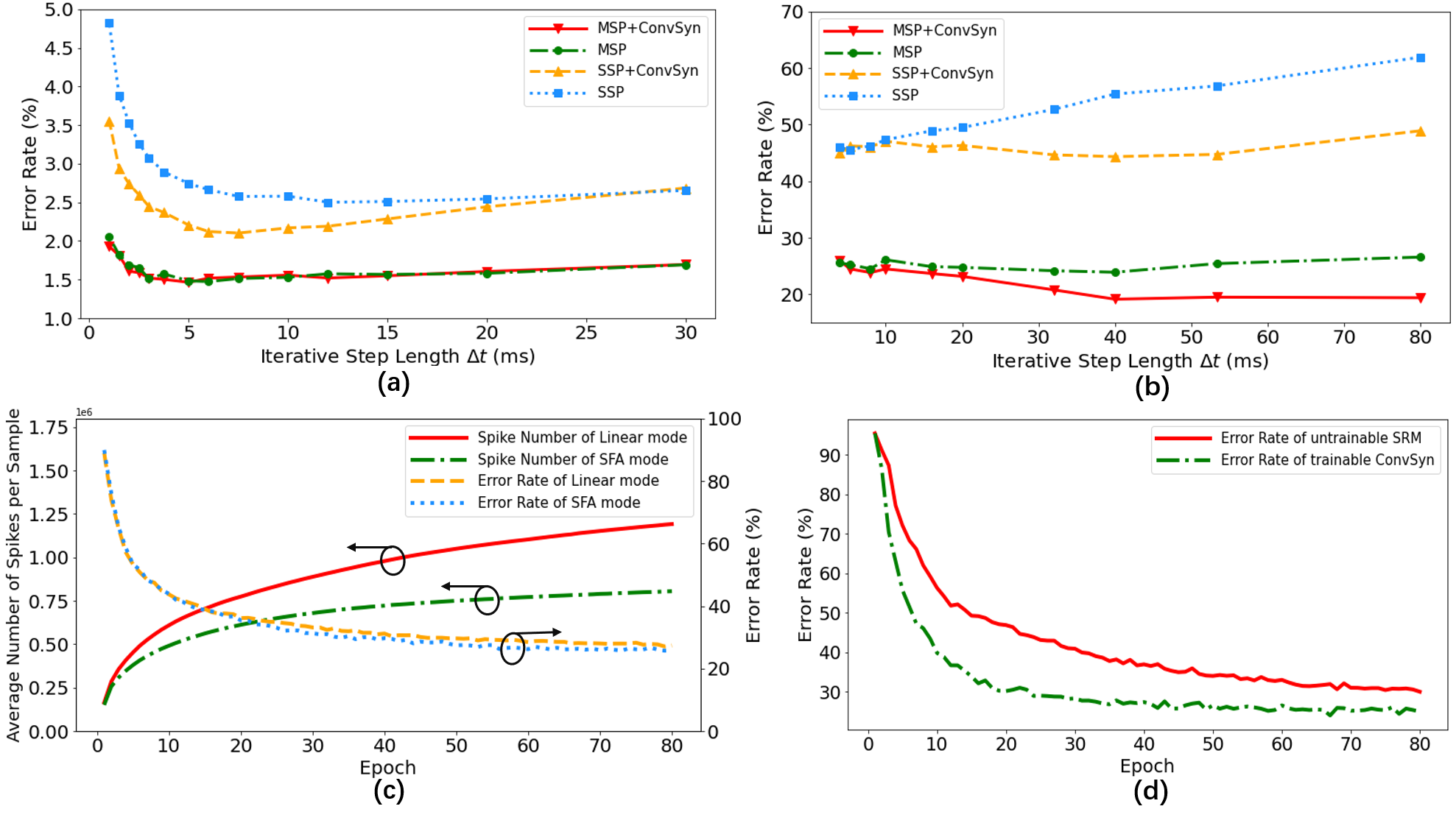}
    \caption{Experimental results. (a) Error rate curves among different iterative step lengths on N-MNIST. (b) Error rate curves among different iterative step lengths on SHD. (c) Control experiment of spike frequencies between SFA and Linear modes on SHD. (d) Control experiment of synaptic plasticity for performance improvement on SHD.}
    \label{fig:results}
\end{figure*}

\subsection{Analysis and Discussion}
To explore  the potentials of the proposed MSP, SFA, and ConvSyn, we carry control experiments on N-MNIST and SHD datasets and discuss the impacts of MAP properties on improving model performance.



\subsubsection{The Impact of Multiplicity on Discrete Iteration}
The selection of minimal iterative step lengths $\Delta t$ influences model performance in the discrete iterative models. For the sake of completeness of the analysis, we analyze this instability in the ablation experiments by building control experiments under MLP architecture with different iterative step lengths, as shown in Figure \ref{fig:results}(a) and Figure \ref{fig:results}(b). The experiments are based on N-MNIST and SHD, respectively, where the unified network structure is 34$\times$34-200-10 on N-MNIST and 700-400-10 on SHD. 
With the additional properties, the error rates of the model have been significantly improved. Compared with benchmark SSP, our MAP-SNN with complementary MSP and ConvSyn reduce error rates by $(1.0\%,2.8\%)$ on N-MNIST, and $(19\%,41\%$) on SHD, which demonstrates the reliability of proposed methods. Furthermore, the model trained with MSP keeps almost constant error rates across different $\Delta t$, supporting that multiplicity alleviates the discretization problem and improves the model stability on time-iteration with arbitrary steps.


\subsubsection{The Impact of Adaptability on Spike Efficiency }
To demonstrate the effectiveness of SFA in spike reduction, we establish a set of controlled experiments on the SHD dataset with the 700-400-10 MLP structure. Figure \ref{fig:results}(c) shows the error rates and spike numbers in the training process of models in both SFA mode and Linear mode. The experimental results show that SFA effectively suppresses spike activities by $1.48\times$ times while slightly improving model accuracy by $1.52\%$. In this case, the reduced signal transmissions helpfully decrease the amount of computation in synapses, which is significant to save the power consumption of neuromorphic hardware based on spike transmissions.

\subsubsection{The Impact of Plasticity on Feature Extraction}
To highlight the importance of plasticity for feature extraction, we set up a control experiment to compare the trainable ConvSyn with the untrainable SRM. As shown in Figure \ref{fig:results}(d), the experiment is set on the SHD with the 700-400-10 MLP structure, showing the changes of model error rate and loss during the training epoch. The experimental results show that the plasticity allows the model to converge faster and reduces the error rate by $(4.2\%,15.6\%)$ during epoch $[10,80]$, demonstrating the advantage of ConvSyn in temporal feature extraction. We conclude that plasticity helps shorten the training process of models and improve the model's performance.

\section{Conclusions}
Inspired by the bionic spike MAP properties, we model spike activities with MSP, SFA, and ConvSyn toward bio-plausible SNNs for better performance. Experimental results confirm the superiority of the proposed model. This work demonstrates the potency of effectively modeling spike activities, revealing a unique perspective for researchers to re-examine the significance of biological facts.

\newpage
\bibliographystyle{named}
\bibliography{paper}

\end{document}


\maketitle

\section{Derivation Details of Formulas}

\subsection{Leaky Integrate-And-Fire as Neural Model}
\subsubsection{LIF Model as differential Equation}
The neural model, Integrate-and-Fire model (LIF) is widely used in SNN algorithms, which approximates real biological neurons and stimulates the accumulation, leakage, and excitation of the membrane potential, using the mathematical model:
\begin{equation}
    \tau\frac{dv}{dt} = - \left\lbrack {v - V_{rest}} \right\rbrack + RI\left( t \right)
\label{eq:lif}
\end{equation}%
Here $\tau$ is the neuron soma's time constant, which equals the product of the capacitance C and resistance R. $I(t)$ is the overall pre-synaptic input current and is accumulated into the membrane potential $v(t)$. $V_{rest}$ is the resting potential of neuron soma. 
When $v(t)$ is in the dynamic range ($V_{rest}<v(t) < V_{threshold}$), the neuron activity follows the Eq.\ref{eq:lif}, accumulating the membrane potential among time. 
Once the $v(t)$ arrives at the potential threshold $V_{threshold}$, the neuron fires a spike and resets the membrane potential $v$ to $V_{rest}$ waiting for accumulation again. The generation of the neuron spike activity $s$ is defined as:
\begin{equation}
\label{eq:s}
s = g(v) = \left\{
\begin{aligned}
0, \quad  v < V_{threshold} \\
1, \quad  v\geq V_{threshold} 
\end{aligned}
\right.
\end{equation}

\subsubsection{Iterable LIF Model}

One approach to process temporal information is to iterate over the time dimension. Iterable neural models applied to time-iteration works similarly to recurrent neural model, which is discretized into minimal iterative step length $\Delta t$.
By setting $V_{rest}=0$ and transforming the first order differentiable equation (Eq.\ref{eq:lif}), the LIF model can be discretized with the iterative step length $\Delta t$ and expressed as:
\begin{equation}
    v(t) = e^{-\frac{\Delta t}{\tau}}\times v(t - \Delta t) + (1-e^{-\frac{\Delta t}{\tau}})\times RI(t)
\label{eq:lif-discrete-1}
\end{equation}%
Equivalently,
\begin{equation}
    v(t) = {\tau_{decay}}\times v(t - 1) + \tilde{I}(t)
\label{eq:lif-discrete-2}
\end{equation}%
Here $\tau_{decay}$ is the time constant describing the leaky activity of LIF model, which equals to $e^{-\frac{\Delta t}{\tau}}$. $\tilde{I}(t)$ is the normalized pre-synaptic input current, which equals to $(1-e^{-\frac{\Delta t}{\tau}})RI(t)$. 
In the discrete simulation, $s(t)$ represents the spike activity during the time interval $(t-\Delta t,t]$, transmitted to the next layer. By unifying the accumulation activity and spike activity, the LIF model under the single-spike pattern for discrete simulation:
\begin{equation}
    v(t) = {\tau_{decay}}\times v(t - 1) \times (1-s(t-1))+ \tilde{I}(t)
\label{eq:lif-discrete-2}
\end{equation}%

\subsection{LIF with Spike Frequency Adaptation}

Under the SFA mechanism, the threshold of LIF will be temporarily raised when a spike occurs to suppress the excitement of dense spike activities. By unifying the accumulation activity and spike activity, the mathematical model of neurons in Spike Frequency Adaptation mode can be described by:
\begin{equation}
     v(t) = \tau_{decay} \times \left\lbrack v\left( {t - \Delta t} \right) - u\left( t - \Delta t \right) \right\rbrack  + \tilde{I}(t)
    \label{eq:SFA}
\end{equation}
Here $v(t)$ is the neuron's membrane potential. $u(t)$ is the consumed membrane potential that produces multiple spike activities. $\tilde{I}(t)$ is the normalized pre-synaptic input current.

The spike activity intensity in SFA is dependent on the membrane potential, modeled with the sum of geometric sequences:
\begin{equation}
    v 
    =\frac{q^{n^{*}} - 1}{q - 1} \times V_{threshold}
\end{equation}

Equivalently,

\begin{equation}
    n^{*} = {\log_{q}\left\lbrack \frac{v}{V_{threshold}}\left( {q - 1} \right) + 1 \right\rbrack}
\end{equation}

\begin{equation}
    s =  n = \left\lfloor n^{*} \right\rfloor
\end{equation}

\begin{figure}[htbp]
    \centering
    \includegraphics[width = 8.5cm]{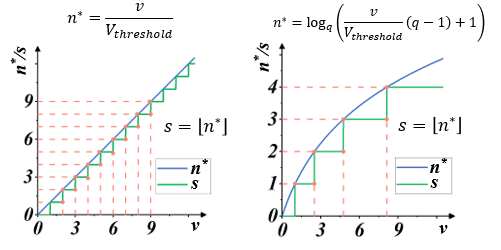}
    \caption{The spike activity versus membrane potential. The blue plot indicates the continuous modeling of the spike intensity, and the green plot indicates the discrete modeling of the spike activities. (a) Spike activities in linear mode. (b) Spike activities in Spike Frequency Adaption mode.}
    \label{fig:SFA}
\end{figure}

Here $n^*$ is the estimated intense spike activity, $q$ is the inhibition coefficient that controls the temporary raising of the threshold. As shown in Figure \ref{fig:SFA}, combined with the Linear mode, the defined SFA mode makes the intensity of spike activity drop exponentially under a given input. Accordingly, the output spike number of neuron activity in SFA takes the integer number $n$:

Based on the spike output $s$, the consumed membrane potential $u(t)$ is calculated by:
\begin{equation}
    u = \sum_{i=1}^{s}{(q^{i}\times V_{threshold})}
    = \frac{q^{s} - 1}{q - 1} \times V_{threshold}
\end{equation}

On this basis, we set $V_{threshold}$, $\tau_{decay}$, $q$ as learnable parameters, which can gradually adjust the membrane potential behavior of neurons in the process of network training. In this case, neurons at the same layer can show different characteristics, improving the ability of model fitting and strengthening the information transmission of the temporal domain.

\subsection{Spike Response Model in Synapses}
\subsubsection{Bio-Electrical Synapses Model}
Spikes are transmitted through synapses in biological neural networks, from the axon terminals to dendrites. With biological electrical synapses, a spike is generated in soma, transmitted through the axon to the synapse, and converted as an electric current to the next neuron's dendrites. One can model this biological behavior using Eq.\ref{eq:neuron_out}.
\begin{align}
    o = f(s,t)
    \label{eq:neuron_out}
\end{align}
Here $s(t)$ is the spike activities, $o(t)$ is the spike response signal, which is transmitted from axon terminal to dendrite over time, $f(\cdot)$ is an activation function that maps spike activities to current intensity through the synapse. The dendrites synthesize the input current intensity from all connected axon terminals and act directly on the membrane potential state of the neuron at some certain time $t$ as Eq.\ref{eq:synapsis}.

\begin{align}
    I_i(t) = \sum_j{w_{ij}o_j(t)}  
    \label{eq:synapsis}
\end{align}
Here $I(t)$ is the overall pre-synaptic input current and is accumulated into the membrane potential, $w_{ij}$ is the synaptic connecting weight from neuron $j$ to neuron $i$.

\begin{figure}[htbp]
    \centering
    \includegraphics[width = 6cm]{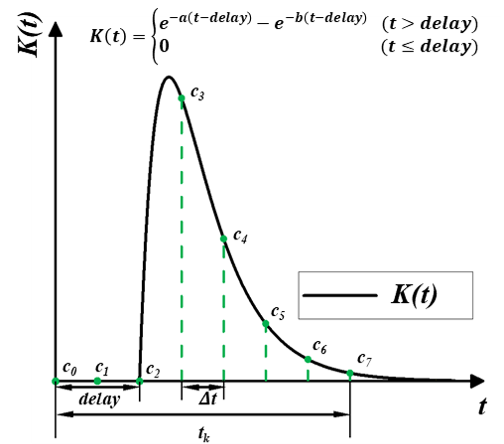}
    \caption{Model the SRM $K(t)$ into 1-D convolution kernel $C_n$. $K(t)$ is modeled with three trainable parameters, with $a$, $b$ for shape, and $delay$ for time delay.}
    \label{fig:SRM}
\end{figure}

\subsubsection{Convolutional Synapse as Spike Response Kernel among Temporal Dimension}

In this paper, to better study the performance of SNN in temporal information extraction, we use the Spike Response Model (SRM) to convert the incoming spikes $s(t)$ into a spike response signal $o(t)$ by convolving $s(t)$ with a spike response kernel $K(\cdot)$:
\begin{equation}
    o(t) = (K\ast s)(t)
\end{equation}

By discretizing the convolutional operation, the spike response signal $o(t)$ can be represented as:


\begin{equation}
    o(t) = \sum\limits_{i=0}^{i\cdot \Delta t \leq t}s(t-i\cdot \Delta t)K(i\cdot \Delta t)
\end{equation}

General expressions for spike response kernel $K(t)$ are 1- and 2-exponential functions such as the following:
\begin{equation}
    K(t) = e^{-t/\tau_d}
\end{equation}
\begin{equation}
    K(t) = e^{-t/\tau_d} - e^{-t/\tau_r}
\end{equation}
Here $\tau_r$ and $\tau_d$ are the rise and decay time constants. The disadvantage of the simple 1-exponential ignores the finite rise time of the synaptic conductance, rising instantaneously from 0 to 1. Hence, the 2-exponential function is used in the proposed model since it contains a finite rise time.

The SRM with axonal delays allows to extract more temporal features. As shown in Figure \ref{fig:SRM}, we adopts the delayed SRM as a convolutional operation with three trainable parameters $a$, $b$, and $delay$ to accelerate the synaptic transmission through the time-iteration under MSP. The definition of convolution synapse is Eq.\ref{eq:conv_syn}:
\begin{equation}
    o\left( t \right) = f(s,t) = {\sum\limits^{i\cdot \Delta t \leq t_k}_{i=0}{s\left( t-i\cdot \Delta t \right) \times C_i}}
    \label{eq:conv_syn}
\end{equation}

\begin{figure*}[htbp]
    \centering
    \includegraphics[width = 16cm]{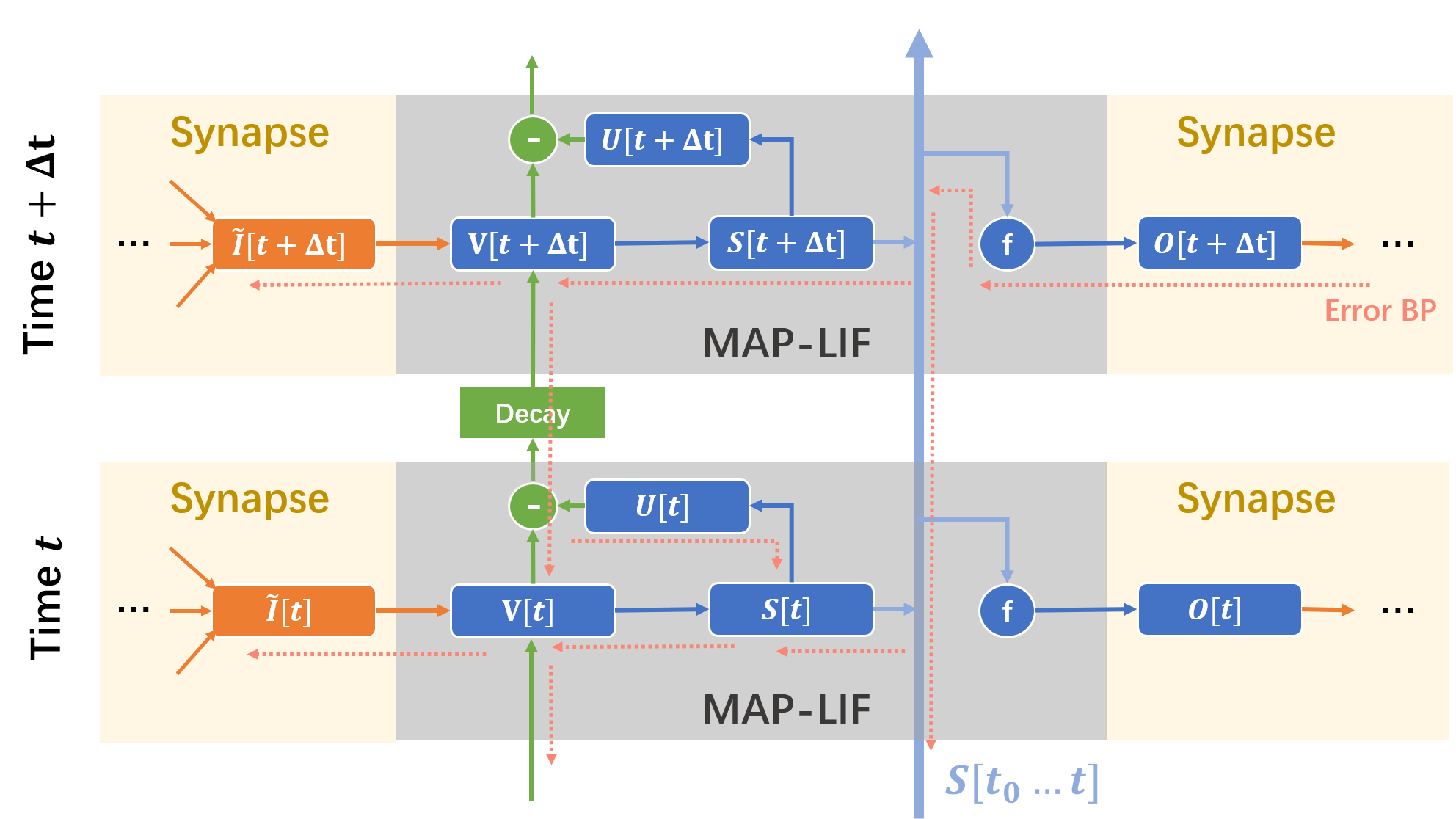}
    \caption{ The spatio-temporal feedforward and backward dataflow in the proposed model. Each box of MAP-LIF indicates a certain spatio-temporal iterative state. The solid arrows indicate spatial and temporal feedforward; the red dotted arrows indicate the error backpropagation correspondingly along with the reverse directions of feedforward.}
    \label{fig:dataflow}
\end{figure*}

Here $C_n$ is the discrete series of spike responses, gradually adjusted during training. $t_k$ is the time-window constant that describes the maximal length of $K(t)$, defined as Eq.\ref{eq:tk} with the minimal iterative step length $\Delta t$ and convolution kernel size. 
\begin{equation}
    t_{k} = (kernel\_size - 1) \times \mathrm{\Delta}t
    \label{eq:tk}
\end{equation}

Then, the discrete self-learning $C_n$ works as delayed spike response kernel $K(t)$. $C_n$ is modeled as:
\begin{equation}
    K^*(t) = e^{-at} - e^{-bt}
    \label{eq:ab}
\end{equation}

\begin{equation}
K(t) = \left\{
\begin{aligned}
K^*(t-delay), \quad  t\geq delay \\
0, \quad  t<delay 
\end{aligned}
\right.
\label{eq:delay}
\end{equation}

\begin{equation}
    C_{i} = K\left( i~ \cdot \mathrm{\Delta}t \right)
    \label{eq:kernel-2}
\end{equation}



Based on the above definition, we can use a simple convolution operation to describe the integration of information in the temporal dimension and adjust the three parameters in the SRM during training.

\subsection{Dendrites with Negative Masking Filter}

In contrast to the biological neural model, the role of dendrites is to receive neurotransmitters from other nerve cells and connect with the axon terminals of the previous cells as synapses. In general, we use multiplication and addition operations to represent the behavior of axon receiving information. Furthermore, to be consistent with the definition of the LIF model, this work set up a negative filter to shield negative values so that the influence of input signal on membrane potential is always non-negative, making the neuron active in only leakage, accumulation, and excitation.

\begin{equation}
    I_{i~} = Filter\left( {\sum_{j}{w_{ij}o_{j}}} \right)
    \label{eq:axon}
\end{equation}

We utilize $Swish$ with $\beta=10$ as the Negative Masking Filter in our model, which makes the negative input tend to be zero in backward:
\begin{equation}
    Filter(x) = Swish(x)= x \cdot Sigmoid(\beta x)
\end{equation}
\begin{equation}
    Sigmoid(z) = \frac{1}{1+e^{-z}}
\end{equation}


\subsection{Dataflow Overall}

Figure \ref{fig:dataflow} shows the spatio-temporal dataflow of neural model in a certain iterative state. We list more specific relationships of each variables below:

\begin{equation}
    \hat{I}_{i}^{n}[t] = Filter(\sum^{L^n}_{i=1}w_{ij}^n O^{n-1}_j[t])
\end{equation}
\begin{equation}
    V_{i}^{n}[t] = \tau_{decay}V_{i}^{n}[t-\Delta t]+\hat{I}_{i}^{n}[t]
\end{equation}
\begin{equation}
    S_{i}^{n}[t] = \left\lfloor  {\log_{q_{i}^{n}}\left\lbrack \frac{V^{n}_{i}[t]}{V^n_{{threshold}_i}}\left( {q_{i}^{n} - 1} \right) + 1 \right\rbrack} \right\rfloor
\end{equation}
\begin{align}
    U_{i}^{n}[t] =& \sum_{k=1}^{S_{i}^{n}[t]}{((q_{i}^{n})^{k}\times V^n_{{threshold}_i})} \nonumber \\
    = &\frac{(q_{i}^{n})^{S_{i}^{n}[t]} - 1}{q_{i}^{n} - 1} \times V^n_{{threshold}_i}
\end{align}
\begin{align}
     O^{n}_j[t] =& f(S_{i}^{n}[t-t_k],...S_{i}^{n}[t-\Delta t], S_{i}^{n}[t]) \nonumber\\
    = &\sum_{d=0}^{d\cdot \Delta t \leq t_k} S_{i}^{n}[t-d\cdot \Delta t] K_i^n(d\cdot \Delta t)
\end{align}%

\begin{equation}
K_i^n(t) =  \left\{
\begin{aligned}
& e^{-a^n_i(t-delay_i^n)} - & e^{-b^n_i(t-delay_i^n)} 
\\
&&\quad  (t\geq delay_i^n) \\
& 0& \\
&&\quad  (t<delay_i^n)
\end{aligned}
\right.
\label{eq:delay}
\end{equation}

\begin{equation}
    Filter(x) = x \cdot \frac{1}{1+e^{-10x}}
\end{equation}

The explanation of formulas is as follows:
\begin{itemize}
    \item The sequences with three indices n, i, t represent states of i-th neuron at the n-th layer and the t-th time point. The variables with two indices n, i represent learnable parameters of i-th neuron at the n-th layer. $L^n$ indicates the number of neurons in the n-th layer. 
    \item $\hat{I}^n_i[t]$ represents the input weighted summation of the output currents from the previous layer and filtered by the function $Filter(\cdot)$
    \item $V^n_i[t]$ represents the neuron's membrane potential.
    \item  $S^n_i[t] \in Z$ represents the number of spikes activities of the neuron within iterative time $(t-\Delta,t]$.
    \item $U^n_i[t]$ represents the consumed mem-potential used for producing spike activities.
    \item $O^n_i[t]$ represents the output currents of the neuron, which is determined by the past spike activities with the convolution operation $f(\cdot)$. The kernel of convolution is defined as $K_i^n(t)$, modeled with $delay_i^n$, $a_i^n$, $b_i^n$.
    \item $V^n_{{threshold}_i}$, $q^n_i$, $a^n_i$, $b^n_i$ and  $delay^n_i$ are all learnable parameters of i-th neuron at the n-th layer, as previously described.

\end{itemize}


